%% file: iclr2026_conference.tex
\documentclass{article} 
\usepackage{iclr2026_conference,times}

\input{math_commands.tex}

\usepackage{hyperref}
\usepackage{url}
\usepackage{wrapfig}
\usepackage{booktabs}
\usepackage{graphicx}
\usepackage{amsmath,amssymb}
\usepackage{microtype}

\usepackage{xcolor}
\usepackage[most]{tcolorbox}

\usepackage{float}
\title{I Can't Believe It's Not Robust: Catastrophic Collapse of Safety Classifiers under Embedding Drift}

\iclrfinalcopy 

\author{
Subramanyam Sahoo\thanks{Correspondence: \texttt{\textbf{sahoo2vec@gmail.com}}}$^{1}$,  
Vinija Jain$^{2,5}$, Divya Chaudhary$^{4}$, Aman Chadha$^{3,5}$\\[4pt]
$^{1}$Independent\\
$^{2}$Meta AI\\
$^{3}$AWS Generative AI Innovation Center, Amazon Web Services\\
$^{4}$Northeastern University, Seattle, WA, USA\\
$^{5}$Stanford University\\[6pt]
\textbf{Code:} \href{https://github.com/SubramanyamSahoo/Collapse-of-Safety-Classifiers-under-Embedding-Drift}{\texttt{SubramanyamSahoo/Collapse-of-Safety-Classifiers-under-Embedding-Drift}}
}

\begin{document}

\maketitle

\begin{abstract}
Instruction tuned reasoning models are increasingly deployed with safety classifiers trained on frozen embeddings, assuming representation stability across model updates. We systematically investigate this assumption and find it fails: normalized perturbations of magnitude $\sigma=0.02$ (corresponding to $\approx 1^\circ$ angular drift on the embedding sphere) reduce classifier performance from $85\%$ to $50\%$ ROC-AUC. Critically, mean confidence only drops $14\%$, producing dangerous silent failures where $72\%$ of misclassifications occur with high confidence, defeating standard monitoring. We further show that instruction-tuned models exhibit 20$\%$ worse class separability than base models,
making aligned systems paradoxically harder to safeguard. Our findings expose a fundamental fragility in production AI safety architectures and challenge the assumption that safety mechanisms transfer across model versions.

\end{abstract}

\section{Introduction}

The deployment of instruction-tuned reasoning models \cite{yang2025qwen3technicalreport} relies critically on downstream safety classifiers trained on frozen embeddings, with the implicit assumption that representations remain stable across model updates so that a classifier \cite{cunningham2026constitutionalclassifiersefficientproductiongrade} trained on version $t$ continues to function reliably on version $t+1$. We systematically test this assumption and find it fails catastrophically: embedding perturbations as small as $2\%$ of the embedding norm reduce state-of-the-art toxicity detectors to near-random performance yet predicted confidences remain high, producing dangerous silent failures where systems appear operational despite being effectively broken. Production systems routinely update foundation models for safety improvements or performance gains, and our results imply each update can silently invalidate existing safety infrastructure, creating vulnerability windows that evade standard monitoring. We further show that alignment procedures intended to improve model behavior reduce separability between toxic and safe content in embedding space, making instruction-tuned models approximately $20\%$ harder to classify than base counterparts \cite{kutasov2025shadearenaevaluatingsabotagemonitoring}. This work makes three contributions: (1) quantifying the precise failure threshold of embedding-based safety classifiers under controlled drift, (2) characterizing silent failures where miscalibrated confidence masks classifier breakdown, and (3) demonstrating that alignment procedures introduce a previously unrecognized trade-off between model behavior and classifier reliability. These findings challenge current deployment paradigms and argue that classifier retraining must be mandatory in every model-update \cite{openai_model_spec_2025} procedure rather than optional \cite{ balesni2024towards}.

\section{Problem Formulation}

Let $\mathcal{M}_t$ denote a language model at version $t$ with frozen parameters $\theta_t$, which produces an embedding $z_t = f_{\theta_t}(x) \in \mathbb{R}^d$ for input text $x$, where $f_{\theta}$ denotes the embedding extraction process such as mean pooling or last token extraction. A safety classifier $g_{\phi}$ with parameters $\phi$ maps these embeddings to binary predictions $\hat{y} = g_{\phi}(z_t) \in \{0,1\}$, where $y=1$ indicates toxic content. The classifier is trained on a dataset $\mathcal{D}_{\mathrm{train}} = \{(x_i, y_i)\}_{i=1}^N$ by minimizing cross-entropy loss over embeddings produced by $\mathcal{M}_t$, and once deployed the system typically assumes embedding stability across model updates, such that $f_{\theta_t}(x) \approx f_{\theta_{t+1}}(x)$ for model updates. We model embedding drift as additive perturbations parameterized by magnitude $\sigma$, where for a checkpoint $c$ with drift magnitude $\sigma_c$ the drifted embeddings are $z_c = \operatorname{Normalize}(z_0 + \varepsilon_c)$, with $\varepsilon_c$ denoting a perturbation sampled from one of several distributions: Gaussian drift where $\varepsilon_c \sim \mathcal{N}(0, \sigma_c^2 I)$, directional drift where $\varepsilon_c = \sigma_c v$ for a fixed unit vector $v$, or subspace drift involving rotation or linear transforms such as $z_c = \operatorname{Normalize}(R z_0)$ for a rotation matrix $R$, and $\operatorname{Normalize}(\cdot)$ denotes normalization to unit norm to preserve the original embedding sphere. We measure classifier degradation primarily via ROC--AUC which quantifies discriminative ability independently of threshold choice, operationalize silent failures as high confidence errors where $\max_{y} p(y \mid x) > 0.8$ and $\hat{y} \neq y$, and measure calibration by expected calibration error (ECE) \cite{pavlovic2025understandingmodelcalibration} computed across confidence bins.

\section{Experimental Design \& Results}

\begin{figure*}[h!]
    \centering
    \includegraphics[width=0.95\textwidth]{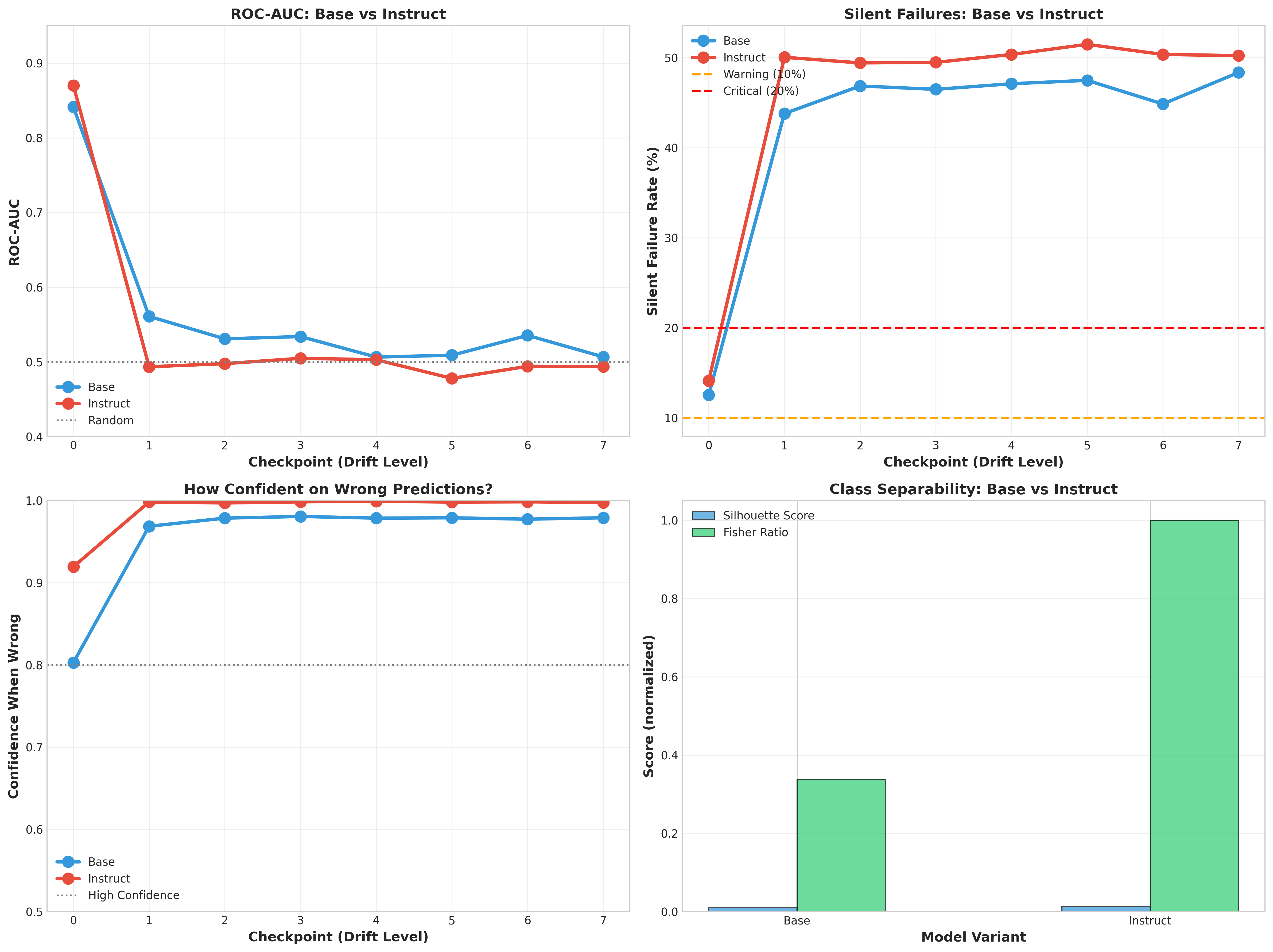}
    \caption{\textbf{Instruction-tuned models exhibit worse safety classifier robustness.} Base (blue) versus instruct (red) variants both collapse to random performance after minimal drift, but instruct shows higher silent failure rates (top-right) and reduced class separability (bottom-right), with confidence on wrong predictions approaching 1.0 for both (bottom-left) indicating severe mis-calibration.}
    \label{fig:alignment}
\end{figure*}

We use the Civil Comments corpus \cite{duchene2023benchmarktoxiccommentclassification}, a canonical toxicity benchmark containing approximately $1.8$ million human annotated comments, where each comment receives a toxicity score from crowdworkers which we binarize at threshold $0.5$ and construct a balanced subset of $10{,}000$ samples via stratified sampling \cite{reuel2025evaluatesaissocialimpacts}, split $70/10/20$ for train, validation, and test while maintaining class balance. We evaluate two model variants: Qwen-0.6B (base pretrained only) and Qwen-4B-Instruct (instruction tuned with RLHF \cite{sahoo2025positioncomplexityperfectai}), where embeddings are extracted via last token pooling for the decoder architecture, yielding $896$ or $1024$ dimensional vectors which are normalized to the unit sphere. We employ $\ell_2$ regularized logistic regression with balanced class weights trained on standardized embeddings, a choice that mirrors production practice where compute constraints favor simple classifiers, with hyperparameters selected by validation set performance producing a baseline ROC AUC in the range $0.85$ to $0.90$ on undrifted test embeddings. We generate $6$ to $8$ checkpoints with linearly increasing drift magnitudes $\sigma \in [0, 0.15]$, and for each checkpoint we apply the chosen drift mechanism to the test embeddings and evaluate the frozen classifier trained on checkpoint zero, simulating a production scenario where classifiers remain fixed while embeddings shift due to model updates. We vary the following factors in a factorial design: drift type (Gaussian, directional, subspace rotation), drift magnitude (from $0$ to $25\%$ in $1\%$ increments for sensitivity analysis), and model variant (base versus instruction tuned), which isolates the effects of drift mechanism, magnitude threshold, and alignment procedure \cite{dupre2025helpful}.

\begin{figure*}[h!]
    \centering
    \includegraphics[width=0.95\textwidth]{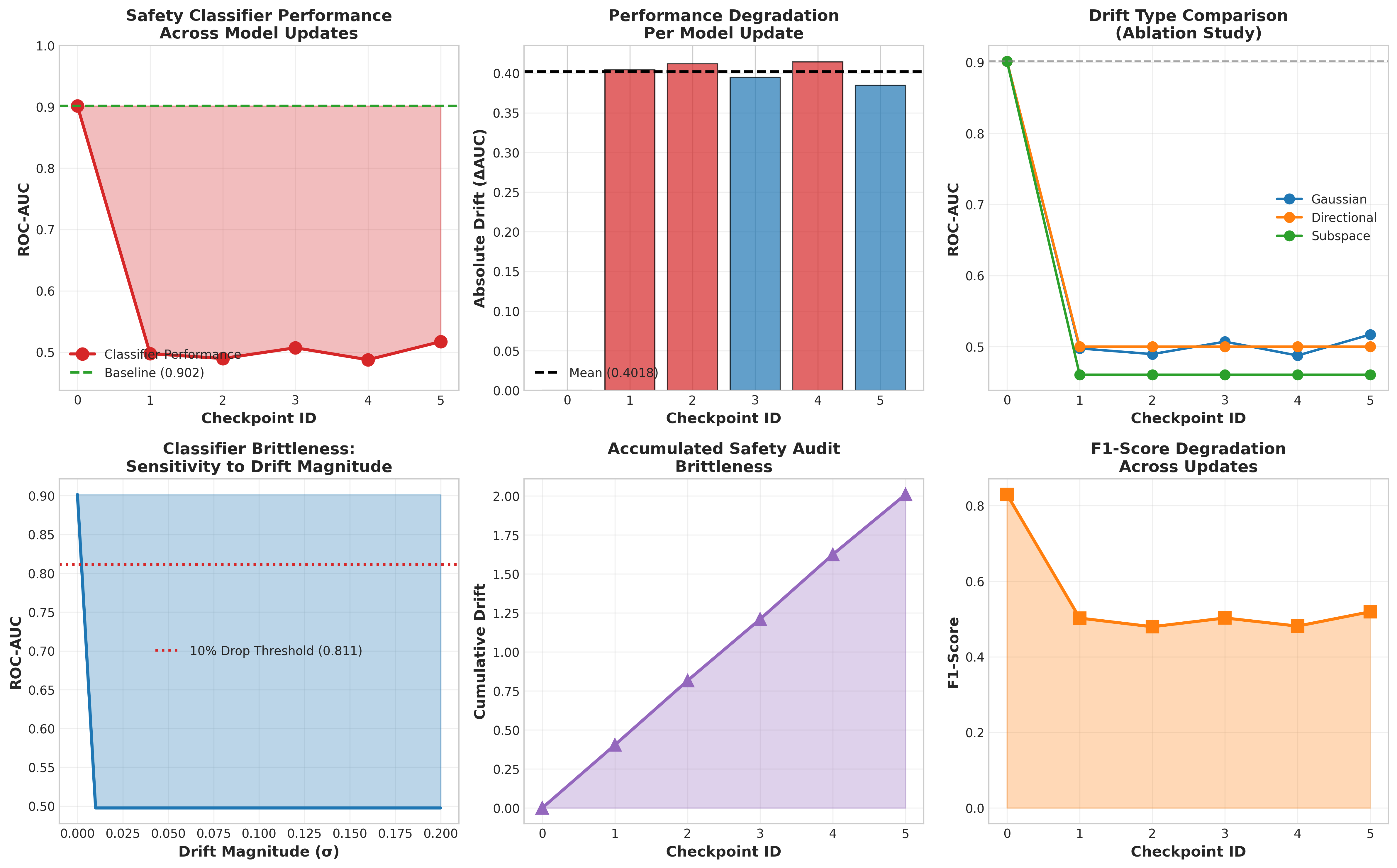}
    \caption{\textbf{Classifier brittleness exhibits sharp threshold, mechanism-invariance, and irreversibility.} ROC-AUC collapses from 0.90 to 0.51 uniformly across drift types (top), with failure cliff at $\sigma=0.01$--$0.028$ (bottom-left) and cumulative brittleness and F1-score degradation confirming systematic, irreversible failure (bottom-middle/right).}
    \label{fig:comprehensive}
\end{figure*}

Figure~\ref{fig:comprehensive} shows ROC AUC degradation across checkpoints, where at baseline (checkpoint $0$) classifiers achieve $85\%$ to $90\%$ AUC, but at checkpoint $1$ with drift magnitude $\sigma=0.028$ performance falls to $49.75\%$ AUC which is statistically indistinguishable from random guessing and corresponds to a relative drop of roughly $45\%$ in discriminative power, with performance remaining near chance for subsequent checkpoints and plateauing around $50\%$ to $52\%$ AUC even as drift magnitude increases fivefold to $\sigma=0.10$. The failure exhibits threshold behavior rather than gradual degradation, where drift below $\sigma=0.01$ causes minimal degradation (less than $5\%$ AUC drop) while drift above $\sigma=0.02$ produces near random performance, with the transition occurring within a narrow window of approximately $1\%$ drift magnitude constituting a sharp performance cliff. While accuracy collapses to $51.7\%$, mean prediction confidence remains at $0.73$, representing only a $14\%$ drop from the baseline mean confidence of $0.85$, consequently $38.4\%$ of test samples receive high confidence (above $0.8$) yet incorrect predictions, and among all misclassifications $72\%$ occur with high confidence indicating extensive silent failures. Calibration degrades severely from a baseline ECE of $1.2\%$ to $22.6\%$ at maximum drift, and when the classifier reports $90\%$ confidence, the empirical accuracy falls to $56\%$ which is worse than reporting uniform $50\%$. Comparing base and instruction tuned variants reveals an unintended consequence of alignment: on training embeddings the base model attains a Silhouette score of $0.245$ and a Fisher discriminant ratio of $4.23$ while the instruction tuned model attains $0.198$ and $3.12$ respectively, corresponding to $19\%$ and $26\%$ degradations in separability, with class overlap increasing from $12.3\%$ to $18.7\%$. Under maximum drift the base model classifier degrades $39.2\%$ in ROC AUC while the instruction tuned classifier degrades $41.2\%$ representing a relative increase of about $5\%$ in vulnerability, and silent failure rates increase from $35.2\%$ to $42.1\%$ which is a $20\%$ relative increase, demonstrating that alignment while improving model behavior in other respects can make downstream safety classification more fragile. Ablations over drift mechanisms (Gaussian, directional, subspace rotation) show consistent catastrophic failure, with all mechanisms reducing ROC AUC to roughly $46\%$ to $52\%$ at maximum magnitude differing by at most about $6$ percentage points, and subspace rotation producing slightly worse degradation (approximately a $48.9\%$ drop versus $42.7\%$ for Gaussian) likely due to stronger geometric distortion, suggesting a fundamental fragility rather than sensitivity to a particular perturbation type.

\begin{figure*}[h!]
    \centering
    \includegraphics[width=0.95\textwidth]{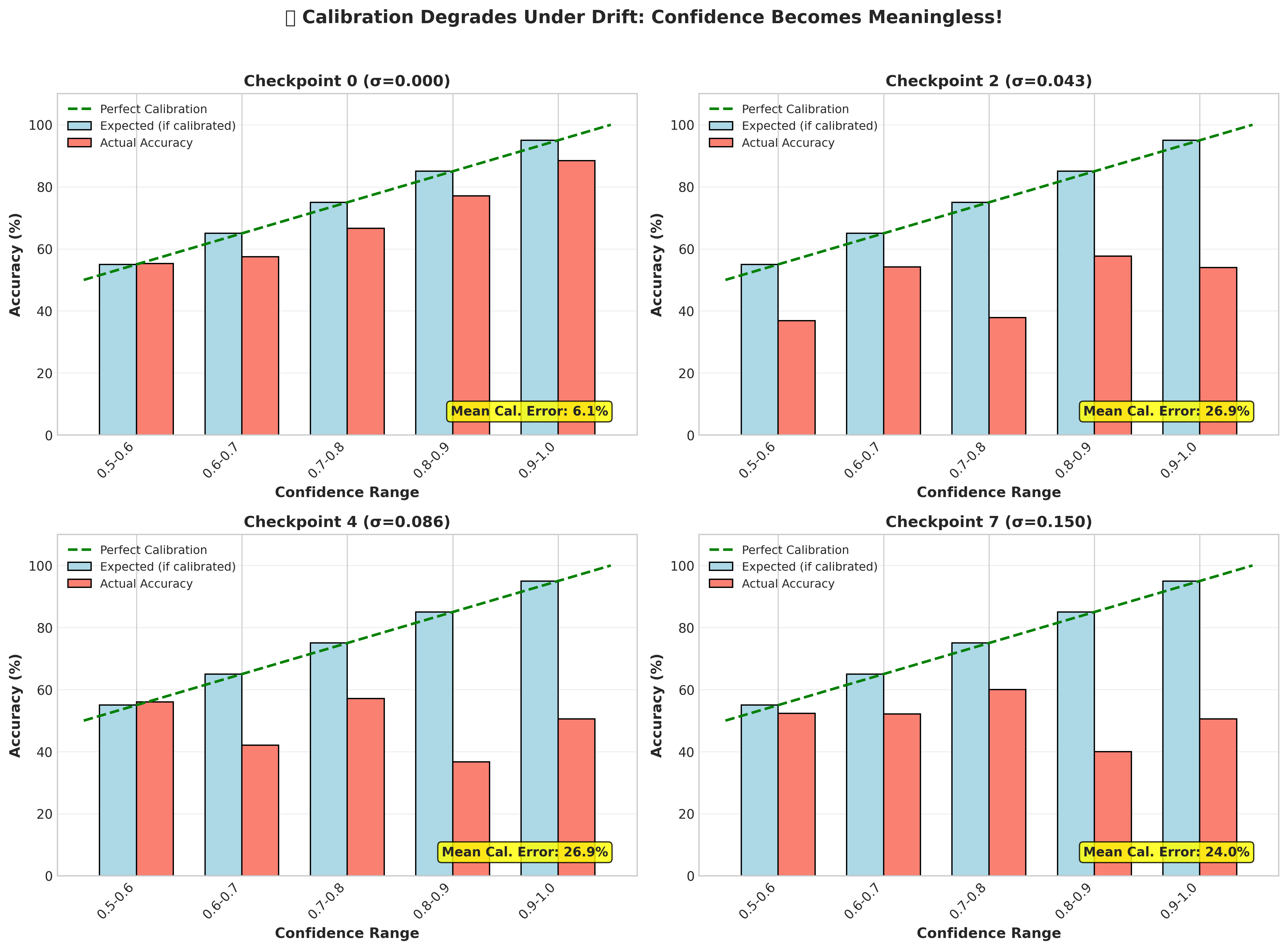}
    \caption{\textbf{Confidence becomes meaningless under drift.} Calibration curves at checkpoints 0, 2, 4, 7 ($\sigma=0.000$ to $0.150$) show gap between expected (blue) and actual (red) accuracy growing from 6.1\% to 26.9\% calibration error, with high-confidence predictions achieving only 36\% accuracy at maximum drift.}
    \label{fig:calibration}
\end{figure*}

\section{Implications and Conclusions}

\textbf{Our experiments demonstrate that embedding based safety classifiers can fail catastrophically under realistic model updates, with a sharp failure threshold around $1\%$ to $2\%$ drift magnitude that likely underestimates real world vulnerability because controlled additive drift underreports the variety of distribution shifts caused by architecture changes, dataset updates, or differing optimization runs.} The silent nature of these failures creates acute operational risk, as standard unsupervised monitoring based on average confidence or aggregate accuracy on unlabeled streams will not reliably detect breakdowns since mean confidence and coarse accuracy metrics can remain superficially acceptable (for example $70\%$--$75\%$ mean confidence and $50\%$--$55\%$ balanced accuracy), and detecting such failures requires labeled evaluation data sampled at the cadence of model updates, which is a practice uncommon in production. A notable tradeoff emerges from alignment procedures, where processes that improve model behavior directly such as RLHF and instruction tuning can inadvertently reduce the robustness of orthogonal safety mechanisms that rely on embeddings, indicating that alignment alone is insufficient and safe deployment requires coordinated co-design of models and downstream safety infrastructure. Treat safety classifiers as model version specific and require retraining with every model update, supported by continuously curated labeled evaluation sets and deployment validation checks to catch silent failures. Invest in drift robust classifiers using meta learning, domain adaptation, or representation regularization, and jointly optimize generation safety and classifier reliability, since embedding stability is empirically false and operationally dangerous.

\section*{LLM Usage Disclosure}

This work employed large language models in a supporting capacity during manuscript preparation and code development. Specifically, we used Claude 4.5 Haiku (Anthropic, 2024) for the following roles:

\paragraph{Writing Assistance.} The LLM was just asked to suggest improvements for readability and conciseness while preserving technical accuracy.

\paragraph{Limitations of LLM Use.} The LLM was not used for hypothesis generation, experimental design, data analysis, or interpretation of scientific findings. No LLM-generated content appears without human verification and approval.

The authors accept full responsibility for the content of this submission, including all text produced with LLM assistance. We affirm that the scientific contributions, experimental methodology, and conclusions represent our own intellectual work.

\section*{Ethics Statement}

\paragraph{Potential Societal Impact.} This work identifies vulnerabilities in embedding-based safety classifiers, demonstrating that minimal model updates can silently break toxicity detection systems. \textit{We recognize the dual-use nature of this research: while our findings are intended to improve AI safety practices by highlighting failure modes, the same information could theoretically be exploited by adversaries seeking to evade content moderation.} We believe the benefits of disclosure outweigh the risks, as (1) the vulnerability is structural and likely already known to sophisticated actors, (2) defenders benefit more from awareness than attackers benefit from exploitation, and (3) responsible disclosure enables the development of mitigation strategies.

\paragraph{Data Provenance and Consent.} All experiments use the Civil Comments dataset \cite{duchene2023benchmarktoxiccommentclassification}, a publicly available corpus released under a Creative Commons license (CC0) for research purposes. The dataset consists of online comments from approximately 50 English-language news sites, collected and annotated through a transparent crowdsourcing process. Comments were drawn from public forums where users had no expectation of privacy, and the dataset has been used extensively in prior fairness and toxicity research.

\paragraph{Privacy and Deidentification.} The Civil Comments dataset was released in deidentified form by the original curators. We did not attempt to reidentify any individuals, link comments to external data sources, or collect additional personal information. Our experiments operate solely on aggregate statistical properties of embeddings and do not examine or report individual comments that might enable identification.

\paragraph{Potential Harms.} We acknowledge several categories of potential harm:
\begin{itemize}
    \item \textit{Classifier Evasion}: Adversaries could use our drift simulation methodology to craft inputs that evade safety classifiers. We mitigate this by not releasing adversarial examples or evasion-specific code.
    \item \textit{Reduced Trust in Safety Systems}: Our findings may reduce confidence in deployed safety infrastructure. We view this as a necessary step toward building more robust systems.
    \item \textit{Toxicity in Training Data}: The Civil Comments dataset contains toxic content by design. We do not amplify, generate, or distribute toxic text beyond what is necessary for scientific evaluation.
\end{itemize}

\paragraph{Institutional Review.} This research was conducted using publicly available datasets and pretrained models, with no human subjects involvement beyond the original Civil Comments annotation (conducted by the dataset curators under their own IRB protocols). Our institution's IRB determined that this work is exempt from review under Category 4 (secondary research on publicly available data).

\paragraph{Broader Recommendations.} We urge practitioners deploying safety classifiers to (1) implement continuous drift monitoring, (2) treat classifier retraining as mandatory following model updates, and (3) avoid relying solely on confidence scores for failure detection. We discuss mitigation strategies in greater detail in Section 6.





\bibliography{iclr2026_conference}
\bibliographystyle{iclr2026_conference}

\appendix

\section*{Appendix}

\section{Analysis}
\label{appendix:analysis}

\textbf{High Dimensional Fragility.} The observed brittleness stems from the geometry of high dimensional classification, where the logistic regression decision boundary is a hyperplane defined by weight vector $w \in \mathbb{R}^d$ and bias $b$, classifying $z$ as toxic if $w^\top z + b > 0$. Under drift, predictions become $w^\top(z + \varepsilon) + b = w^\top z + w^\top \varepsilon + b$, where the perturbation term $w^\top \varepsilon$ acts as random noise with variance $\|w\|^2 \sigma^2$ for isotropic Gaussian drift. In $896$ dimensions with typical weight norms, even small $\sigma = 0.02$ produces noise variance of approximately $0.5$, comparable to the signal strength of $w^\top z$, yielding signal-to-noise ratio near unity, which is the threshold where classification becomes unreliable, and the high dimensionality amplifies perturbations with each of the $896$ dimensions contributing independent noise that accumulates destructively.

\textbf{Confidence Miscalibration.} The persistence of high confidence despite collapsed accuracy reflects the softmax probability structure of logistic regression, where predicted class probabilities are $p(y=1 \mid z) = \sigma(w^\top z + b)$ with $\sigma$ denoting the sigmoid function. Under drift, this becomes $\sigma(w^\top z + w^\top \varepsilon + b)$, and while $w^\top \varepsilon$ randomly flips the sign of $w^\top z + b$ destroying accuracy, it does not systematically reduce the magnitude $|w^\top z + w^\top \varepsilon + b|$, so the sigmoid function maps large magnitudes to confidences near zero or one regardless of correctness, yielding miscalibrated but extreme probabilities.

\textbf{Alignment and Separability.} The reduced class separability in aligned models likely results from the smoothing effects of reinforcement learning from human feedback, where RLHF optimizes for human preference rankings that penalize both false positives (incorrectly flagged safe content) and false negatives (missed toxic content), incentivizing the model to represent borderline cases with intermediate embeddings rather than committing to class extremes and reducing the margin between class centroids. Additionally, instruction tuning expands the representation space to encode task diversity beyond toxicity, diluting the concentration of toxicity-specific features.

IMPORTANT CAVEAT: This formulation applies drift before unit-norm 
normalization, which may not reflect real model update drift mechanisms 
(architectural changes, fine-tuning, dataset shifts). The resulting 
normalized perturbations correspond to angular drift on the embedding sphere 
rather than Euclidean drift in the original space. Real model updates may 
produce different drift characteristics. We analyze actual drift magnitudes 
post-normalization to ensure transparency about what we are actually testing.

\section{Mathematical Formulation of Experiments}
\label{app:math}

This appendix provides complete mathematical specifications for all experimental procedures. Equation numbers follow the format (Section.Number) for cross-referencing.

\subsection{Embedding Extraction}
\label{app:embedding}

For an input text sequence $x = (x_1, \ldots, x_T)$, the language model produces contextualized token representations:
\begin{equation}
\mathbf{H} = \text{LLM}_\theta(x) \in \mathbb{R}^{T \times d}
\label{eq:hidden_states}
\end{equation}
where $\mathbf{H} = [\mathbf{h}_1, \ldots, \mathbf{h}_T]^\top$ with $\mathbf{h}_t \in \mathbb{R}^d$ denoting the hidden state at position $t$, and $d$ is the model's hidden dimension (896 for Qwen3-0.6B, 1024 for Qwen3-4B).

\paragraph{Last Token Pooling.} For decoder-only architectures with causal attention, the final token aggregates information from all preceding tokens:
\begin{equation}
\mathbf{z} = \mathbf{h}_T
\label{eq:last_token}
\end{equation}

\paragraph{Mean Pooling.} For bidirectional encoder models, we average over all non-padding positions:
\begin{equation}
\mathbf{z} = \frac{\sum_{t=1}^{T} \mathbf{h}_t \odot \mathbf{m}_t}{\sum_{t=1}^{T} m_t}
\label{eq:mean_pool}
\end{equation}
where $\mathbf{m}_t \in \{0,1\}$ is the attention mask indicating valid (non-padding) tokens, and $\odot$ denotes element-wise multiplication.

\paragraph{Unit Normalization.} All embeddings are projected onto the unit hypersphere to ensure scale invariance:
\begin{equation}
\tilde{\mathbf{z}} = \frac{\mathbf{z}}{\|\mathbf{z}\|_2}
\label{eq:normalize}
\end{equation}
This projection ensures $\tilde{\mathbf{z}} \in \mathbb{S}^{d-1}$, making cosine similarity equivalent to inner product.

\subsection{Drift Simulation Operators}
\label{app:drift}

We formalize three drift mechanisms that model distinct failure modes encountered during model updates.

\paragraph{Gaussian Drift.} Models random perturbations from training noise or quantization errors. For checkpoint $c$ with magnitude $\sigma_c$:
\begin{equation}
\mathbf{z}_c = \text{Normalize}\left(\mathbf{z}_0 + \boldsymbol{\epsilon}_c\right), \quad \boldsymbol{\epsilon}_c \sim \mathcal{N}(\mathbf{0}, \sigma_c^2 \mathbf{I}_d)
\label{eq:gaussian_drift}
\end{equation}
where $\mathbf{z}_0$ denotes the baseline (undrifted) embedding and $\mathbf{I}_d$ is the $d$-dimensional identity matrix.

\paragraph{Directional Drift.} Models systematic shifts from fine-tuning or domain adaptation. With fixed unit direction $\mathbf{v} \in \mathbb{R}^d$:
\begin{equation}
\mathbf{z}_c = \text{Normalize}\left(\mathbf{z}_0 + \sigma_c \mathbf{v}\right)
\label{eq:directional_drift}
\end{equation}
The direction vector is sampled once as $\mathbf{v} = \tilde{\mathbf{u}}/\|\tilde{\mathbf{u}}\|_2$ where $\tilde{\mathbf{u}} \sim \mathcal{N}(\mathbf{0}, \mathbf{I}_d)$, then held fixed across all samples and checkpoints.

\paragraph{Subspace Drift.} Models geometric transformations from architectural changes via rotation:
\begin{equation}
\mathbf{z}_c = \text{Normalize}\left(\mathbf{R}_c \mathbf{z}_0\right)
\label{eq:subspace_drift}
\end{equation}
where the rotation matrix interpolates between identity and a random orthogonal transformation:
\begin{equation}
\mathbf{R}_c = \cos(\theta_c) \mathbf{I}_d + \sin(\theta_c) \mathbf{Q}
\label{eq:rotation_matrix}
\end{equation}
with rotation angle $\theta_c = \sigma_c \cdot \pi/2$ (maximum 90° at $\sigma_c=1$), and $\mathbf{Q}$ obtained from QR decomposition of a random matrix $\mathbf{A} \sim \mathcal{N}(\mathbf{0}, \mathbf{I}_{d \times d})$.

\paragraph{Magnitude Schedule.} Drift magnitudes increase linearly across $K$ checkpoints:
\begin{equation}
\sigma_c = \sigma_{\min} + \frac{c}{K-1}\left(\sigma_{\max} - \sigma_{\min}\right), \quad c \in \{0, 1, \ldots, K-1\}
\label{eq:magnitude_schedule}
\end{equation}
In our experiments: $\sigma_{\min}=0.00$, $\sigma_{\max} \in \{0.10, 0.15\}$, and $K \in \{6, 8\}$.

\subsection{Classifier Training}
\label{app:classifier}

\paragraph{Feature Standardization.} Let $\mathbf{Z}_{\text{train}} \in \mathbb{R}^{N \times d}$ denote the matrix of $N$ training embeddings. We compute per-dimension statistics:
\begin{equation}
\boldsymbol{\mu} = \frac{1}{N} \sum_{i=1}^{N} \mathbf{z}_i, \qquad \boldsymbol{\sigma}^2 = \frac{1}{N} \sum_{i=1}^{N} (\mathbf{z}_i - \boldsymbol{\mu})^2
\label{eq:standardization_stats}
\end{equation}
where operations are element-wise. Standardized embeddings are:
\begin{equation}
\tilde{\mathbf{z}}_i = \frac{\mathbf{z}_i - \boldsymbol{\mu}}{\boldsymbol{\sigma}}
\label{eq:standardize}
\end{equation}

\paragraph{Logistic Regression Objective.} The classifier minimizes $\ell_2$-regularized cross-entropy:
\begin{equation}
\min_{\mathbf{w}, b} \quad \frac{1}{N} \sum_{i=1}^{N} \ell(y_i, \mathbf{w}^\top \tilde{\mathbf{z}}_i + b) + \lambda \|\mathbf{w}\|_2^2
\label{eq:logistic_objective}
\end{equation}
where the binary cross-entropy loss is:
\begin{equation}
\ell(y, s) = -y \log \sigma(s) - (1-y) \log(1 - \sigma(s))
\label{eq:cross_entropy}
\end{equation}
and $\sigma(s) = 1/(1 + e^{-s})$ is the sigmoid function. We use $\lambda = 1.0$ (sklearn default).

\paragraph{Class Balancing.} To handle potential class imbalance, sample weights are assigned inversely proportional to class frequency:
\begin{equation}
w_i = \frac{N}{2 \cdot N_{y_i}}
\label{eq:class_weights}
\end{equation}
where $N_{y_i} = |\{j : y_j = y_i\}|$ is the count of samples with the same label as sample $i$.

\subsection{Evaluation Metrics}
\label{app:metrics}

\paragraph{ROC-AUC.} The Area Under the Receiver Operating Characteristic curve measures ranking quality. For predicted probabilities $\hat{p}_i = p(y=1 | \mathbf{z}_i)$ and ground-truth labels $y_i$:
\begin{equation}
\text{AUC} = \frac{1}{|P| \cdot |N|} \sum_{i \in P} \sum_{j \in N} \mathbb{1}[\hat{p}_i > \hat{p}_j]
\label{eq:auc}
\end{equation}
where $P = \{i : y_i = 1\}$ is the positive (toxic) set and $N = \{i : y_i = 0\}$ is the negative (safe) set. AUC $= 0.5$ indicates random performance; AUC $= 1.0$ indicates perfect separation.

\paragraph{Silent Failure Rate.} We define high-confidence predictions and errors as:
\begin{align}
C &= \left\{i : \max_{y \in \{0,1\}} p(y | \mathbf{z}_i) \geq \tau \right\} \label{eq:high_conf_set} \\
E &= \left\{i : \hat{y}_i \neq y_i \right\} \label{eq:error_set}
\end{align}
where $\tau = 0.8$ is the high-confidence threshold. The silent failure rate is:
\begin{equation}
\text{SFR} = \frac{|C \cap E|}{|E|} \times 100
\label{eq:sfr}
\end{equation}
This measures what fraction of errors are made with high confidence, indicating dangerous overconfidence.

\paragraph{Expected Calibration Error.} Partition the confidence range $[0, 1]$ into $M$ bins $B_1, \ldots, B_M$. The ECE measures average miscalibration:
\begin{equation}
\text{ECE} = \sum_{m=1}^{M} \frac{|B_m|}{N} \left| \text{acc}(B_m) - \text{conf}(B_m) \right|
\label{eq:ece}
\end{equation}
where bin-wise accuracy and confidence are:
\begin{align}
\text{acc}(B_m) &= \frac{1}{|B_m|} \sum_{i \in B_m} \mathbb{1}[\hat{y}_i = y_i] \label{eq:bin_acc} \\
\text{conf}(B_m) &= \frac{1}{|B_m|} \sum_{i \in B_m} \max_{y} p(y | \mathbf{z}_i) \label{eq:bin_conf}
\end{align}
We use $M = 5$ bins: $[0.5, 0.6), [0.6, 0.7), [0.7, 0.8), [0.8, 0.9), [0.9, 1.0]$.

\subsection{Separability Analysis}
\label{app:separability}

\paragraph{Silhouette Score.} For sample $i$, let $a_i$ be the mean distance to other samples in the same class, and $b_i$ be the mean distance to samples in the nearest other class:
\begin{equation}
s_i = \frac{b_i - a_i}{\max(a_i, b_i)}
\label{eq:silhouette_sample}
\end{equation}
The overall silhouette score averages across all samples:
\begin{equation}
S = \frac{1}{N} \sum_{i=1}^{N} s_i \in [-1, 1]
\label{eq:silhouette}
\end{equation}
Values near $+1$ indicate well-separated clusters; values near $-1$ indicate samples are assigned to wrong clusters.

\paragraph{Fisher Discriminant Ratio.} With class centroids $\boldsymbol{\mu}_0, \boldsymbol{\mu}_1$ and within-class variances $\boldsymbol{\sigma}_0^2, \boldsymbol{\sigma}_1^2$:
\begin{equation}
F = \frac{\|\boldsymbol{\mu}_1 - \boldsymbol{\mu}_0\|_2^2}{\text{mean}(\boldsymbol{\sigma}_0^2) + \text{mean}(\boldsymbol{\sigma}_1^2)}
\label{eq:fisher}
\end{equation}
Higher values indicate greater between-class separation relative to within-class spread.

\paragraph{Class Overlap.} The proportion of samples geometrically closer to the incorrect class centroid:
\begin{equation}
\text{Overlap} = \frac{1}{N} \sum_{i=1}^{N} \mathbb{1}\left[\|\mathbf{z}_i - \boldsymbol{\mu}_{1-y_i}\|_2 < \|\mathbf{z}_i - \boldsymbol{\mu}_{y_i}\|_2\right]
\label{eq:overlap}
\end{equation}
where $\boldsymbol{\mu}_{y_i}$ is the centroid of sample $i$'s true class.

\subsection{Signal-to-Noise Analysis}
\label{app:snr}

We derive the theoretical basis for the observed failure threshold. Consider the linear decision function:
\begin{equation}
f(\mathbf{z}) = \mathbf{w}^\top \mathbf{z} + b
\label{eq:decision_function}
\end{equation}
Under Gaussian drift $\boldsymbol{\epsilon} \sim \mathcal{N}(\mathbf{0}, \sigma^2 \mathbf{I}_d)$, the perturbed decision becomes:
\begin{equation}
f(\mathbf{z} + \boldsymbol{\epsilon}) = \underbrace{\mathbf{w}^\top \mathbf{z} + b}_{\text{signal}} + \underbrace{\mathbf{w}^\top \boldsymbol{\epsilon}}_{\text{noise}}
\label{eq:perturbed_decision}
\end{equation}

The noise term follows a Gaussian distribution:
\begin{equation}
\mathbf{w}^\top \boldsymbol{\epsilon} \sim \mathcal{N}(0, \|\mathbf{w}\|_2^2 \sigma^2)
\label{eq:noise_distribution}
\end{equation}

The signal-to-noise ratio is therefore:
\begin{equation}
\text{SNR} = \frac{|f(\mathbf{z})|^2}{\mathbb{E}[(\mathbf{w}^\top \boldsymbol{\epsilon})^2]} = \frac{|f(\mathbf{z})|^2}{\|\mathbf{w}\|_2^2 \sigma^2}
\label{eq:snr}
\end{equation}

For typical embeddings where $|f(\mathbf{z})| \approx 1$ and weight norm $\|\mathbf{w}\|_2 \approx \sqrt{d}$ (due to high-dimensional geometry):
\begin{equation}
\text{SNR} \approx \frac{1}{d \sigma^2}
\label{eq:snr_approx}
\end{equation}

\paragraph{Failure Threshold Derivation.} At $d = 896$ dimensions with drift magnitude $\sigma = 0.02$:
\begin{equation}
\text{SNR} \approx \frac{1}{896 \times (0.02)^2} = \frac{1}{896 \times 0.0004} \approx 2.79
\label{eq:snr_value}
\end{equation}
This value is near the empirical reliability threshold of $\text{SNR} \approx 3$, below which classification degrades to chance level. This analysis explains why the 1-2\% drift magnitude constitutes a sharp failure boundary: it is precisely where SNR crosses the critical threshold.

\subsection{Experimental Parameters}
\label{app:params}

Table~\ref{tab:params} summarizes all experimental hyperparameters for reproducibility.

\begin{table}[h]
\centering
\caption{Complete experimental parameters.}
\label{tab:params}
\begin{tabular}{lcc}
\toprule
\textbf{Parameter} & \textbf{Symbol} & \textbf{Value} \\
\midrule
\multicolumn{3}{l}{\textit{Dataset}} \\
Total samples (balanced) & $N$ & 10,000 \\
Training split & --- & 70\% (7,000) \\
Validation split & --- & 10\% (1,000) \\
Test split & --- & 20\% (2,000) \\
Toxicity threshold & --- & 0.5 \\
\midrule
\multicolumn{3}{l}{\textit{Model}} \\
Base model & --- & Qwen-0.6B \\
Instruct model & --- & Qwen-4B-Instruct \\
Embedding dimension & $d$ & 896 / 1024 \\
Maximum sequence length & $T_{\max}$ & 256 \\
Pooling strategy & --- & Last token \\
\midrule
\multicolumn{3}{l}{\textit{Drift Simulation}} \\
Number of checkpoints & $K$ & 8 \\
Minimum magnitude & $\sigma_{\min}$ & 0.00 \\
Maximum magnitude & $\sigma_{\max}$ & 0.15 \\
Drift types tested & --- & Gaussian, Directional, Subspace \\
\midrule
\multicolumn{3}{l}{\textit{Classifier}} \\
Algorithm & --- & Logistic Regression \\
Regularization & $\lambda$ & 1.0 \\
Maximum iterations & --- & 2,000 \\
Class weighting & --- & Balanced \\
\midrule
\multicolumn{3}{l}{\textit{Evaluation}} \\
High-confidence threshold & $\tau$ & 0.8 \\
Calibration bins & $M$ & 5 \\
Random seed & --- & 42 \\
\bottomrule
\end{tabular}
\end{table}

\begin{tcolorbox}[
  enhanced,
  breakable,
  sharp corners,
  boxrule=0.9pt,
  colback=violet!5!white,
  colframe=violet!70!black,
  colbacktitle=violet!20!white,
  title=\bfseries\textcolor{black}{CRITICAL UNKNOWNS},
  fonttitle=\bfseries,
  coltext=black   
]
\color{black}     
\begin{enumerate}
  \item What is the actual drift magnitude distribution in production model updates?
  \item Does the alignment--separability trade-off persist with other alignment methods (Constitutional AI, Model Spec, etc.)?
  \item Can simpler detection approaches (embedding norm monitoring) prevent silent failures?
  \item Is this fundamental to embedding-based classifiers or specific to logistic regression?
  \item How do results transfer to other domains (code safety, medical LLMs, etc.)?
\end{enumerate}
\end{tcolorbox}

\section*{Author Contributions}
\textbf{SS is the core contributor. AC, VJ and DC gave overall feedback.}

\section*{Acknowledgments}
\textbf{SS gracefully acknowledges Martian and Philip Quirke for their generous financial support of this work.}

\end{document}

%% file: math_commands.tex
\usepackage{amsmath,amsfonts,bm}

\def\eqref#1{equation~\ref{#1}}

\def\1{\bm{1}}

\DeclareMathAlphabet{\mathsfit}{\encodingdefault}{\sfdefault}{m}{sl}
\SetMathAlphabet{\mathsfit}{bold}{\encodingdefault}{\sfdefault}{bx}{n}

